\documentclass[letterpaper, 10 pt, journal, twoside]{IEEEtran}

\IEEEoverridecommandlockouts                              

\usepackage{amsmath} 
\usepackage{amssymb}  
\usepackage{bm, balance}
\usepackage{algorithm,algpseudocode}
\usepackage[small]{caption}
\usepackage{multicol}
\usepackage{subcaption}
\usepackage{graphicx}
\usepackage{xcolor,mathrsfs,array}
\usepackage[export]{adjustbox}
\usepackage[utf8]{inputenc}
\usepackage[T1]{fontenc}
\usepackage{cite}
\usepackage{flushend}
\usepackage{lipsum}
\usepackage{svg}
\usepackage{booktabs}
\usepackage{multirow}

\usepackage{cuted}
\usepackage{capt-of}
\makeatletter
\def\Huge{\@setfontsize\Huge{16}{20pt}}  
\makeatother

\usepackage[linkbordercolor={1 1 1},citebordercolor={1 1
  1},urlbordercolor={0.0 0.0
  0.0},urlcolor=blue,colorlinks=true,linkcolor=black,citecolor=black]{hyperref}
\usepackage{url}

\DeclareUnicodeCharacter{2217}{*}

\usepackage{shortcuts} 

\usepackage{pifont}
\newcommand{\cmark}{\ding{51}}%
\newcommand{\xmark}{\ding{55}}%
\newcommand{\morlax}{\texttt{MORLAX}}
\newcommand{\moplayground}{MO-Playground}
\newcommand{\speedup}{26-271x}

\usepackage{etoolbox}
\newtoggle{showrevisions}                                             
\togglefalse{showrevisions}
\newcommand{\revisedii}[1]{\iftoggle{showrevisions}{\textcolor{red}{#1}}{#1}}                                 
\newcommand{\revisediii}[1]{\iftoggle{showrevisions}{\textcolor{blue}{#1}}{#1}}                               
\newcommand{\revised}[1]{\iftoggle{showrevisions}{\textcolor{cyan}{#1}}{#1}} 

\begin{document}


\title{{\fontsize{16pt}{20pt}\selectfont \underline{MO}-Playground$^1$: 

Massively Parallelized \underline{M}ulti-\underline{O}bjective Reinforcement Learning for Robotics}}


\author{Neil Janwani$^{1\dagger}$, Ellen Novoseller$^{2}$, Vernon J. Lawhern$^{2}$, Maegan Tucker$^{1}$
\thanks{Manuscript recieved: February 27, 2026. Accepted: April 26, 2026.}
\thanks{
This paper was recommended for publication by
Editor Wei Pan upon evaluation of the Associate Editor and Reviewers comments.}
\thanks{
This work was supported by DEVCOM Army Research Laboratory under Cooperative Agreement Number W911NF-19-S-0001 and the Institute of Robotics and Intelligent Machines at Georgia Tech.}
\thanks{$\dagger$Corresponding author: \texttt{\{njanwani, mtucker\}@gatech.edu}, }
\thanks{$^1$Georgia Institute of Technology, Atlanta, GA USA}
\thanks{$^2$DEVCOM Army Research Laboratory, MD, USA}
\thanks{Digital Object Identifier (DOI): see top of this page.}
}

\markboth{IEEE Robotics and Automation Letters. Preprint Version. Accepted MAY, 2026}
{Janwani \MakeLowercase{\textit{et al.}}: MO-Playground}

\maketitle

\begin{strip}
    \vspace{-3.6cm}
    \centering
    \includegraphics[width=\linewidth]{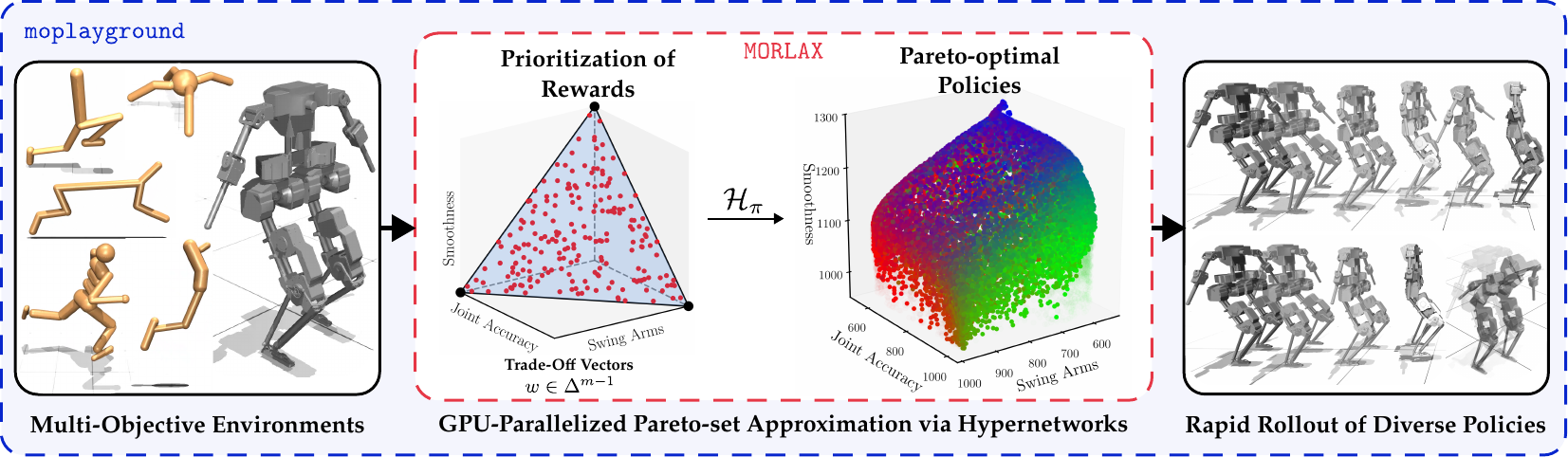}
    \captionof{figure}{\moplayground~provides a suite of multi-objective environments, a framework for easily creating custom environments, 
    and fast Pareto set approximation via GPU-accelerated massive parallelization and parameter-efficient hypernetworks.}
    \label{fig:teaser}
    \vspace{-5pt} 
\end{strip}

\begin{abstract}

Multi-objective reinforcement learning (MORL) is a powerful tool to learn Pareto-optimal policy families across conflicting objectives. However, unlike traditional RL algorithms, existing MORL algorithms do not effectively leverage large-scale parallelization to concurrently simulate thousands of environments, \revised{thus facing} vastly increased computation time. Ultimately, this has limited \revised{the application of MORL} towards complex multi-objective robotics problems.
To address these challenges, we present 1) \morlax, a new GPU-native, fast MORL algorithm, and 2) \moplayground, a \texttt{pip}-installable playground of GPU-accelerated multi-objective environments. Together, \morlax~and \moplayground~approximate Pareto sets within minutes, offering \speedup~speed-ups compared to legacy CPU-based approaches and up to 19x speed-ups over prior GPU-based approaches whilst \revised{learning} superior Pareto front hypervolumes. We demonstrate \moplayground's versatility by implementing a custom BRUCE humanoid robot environment and learning Pareto-optimal locomotion policies across 6 \revisedii{practical} objectives \revisedii{in simulation}, such as smoothness, efficiency and arm swinging. 
Our toolbox is available at 
\url{https://dynamicmobility.github.io/moplayground/}.
\end{abstract}


\IEEEpeerreviewmaketitle

\vspace{-4mm}
\section{Introduction}

Reinforcement learning (RL) has established itself as an effective paradigm for synthesizing complex robotic controllers by enabling agents to learn directly from interactions with their environment \cite{hwangbo2019learning, rudin2022learning, liao2025beyondmimic}. However, this \revised{framework} notoriously relies on crafting a \revised{fixed}, single-objective, scalar reward function that trades off between conflicting objectives such as energy efficiency and tracking accuracy. This reward shaping process is time-consuming, requires domain \revised{expertise, and necessitates a pre-defined, fixed weighting over objectives} 
at training time, presenting a major barrier to real-world deployment \cite{hayes2022practical}.

\revised{One prominent approach for RL with unknown rewards is to learn from human labels, such as}
demonstrations, pairwise preferences, or ratings \cite{ravichandar2020recentlfd, janwani2025navigait, liao2025beyondmimic, tucker2020preference, li2021roial, white2024rating}. Yet, these methods \revised{infer} a single reward function, and do not learn to intelligently trade off between \revised{multiple} conflicting objectives.
This lack of on-demand flexibility is especially problematic for real-world robotic systems, \revised{whose optimal trade-offs across objectives vary by scenario.} For instance, human-robot systems such as lower-limb exoskeletons or assistive driving require tunability for user-specific personalization \cite{tucker2020preference, basu2017you, janwani2024learning}. 


Multi-Objective Reinforcement Learning (MORL) provides a principled framework to address this gap by replacing the scalar reward with a vectorized one,\revised{ where each component represents a different objective.}
With this vectorized reward, MORL enables practitioners to flexibly tune robot behavior post-training by finding and exploring the \textit{Pareto set}, a collection of high quality solutions that are optimal \revised{for different scenarios}, capturing critical trade-offs among objectives. 
To this end, a wide variety of MORL algorithms have been developed to approximate or explore the Pareto set efficiently \cite{hayes2022practical, felten_toolkit_2023}, including evolutionary approaches \cite{xu2020prediction}, context-conditioned policy exploration \cite{alegre2025amor, basaklar2023pd}, and hypernetwork-based approaches \cite{hypermorl, liu2025paretosetlearningmultiobjective}. 
\revised{Despite these methodological advances}, modern MORL algorithms still require long training times--sometimes \revised{several} days \cite{alegre2025amor}.

Meanwhile, single-objective RL has benefited from hardware-accelerated differentiable physics and vectorized computation frameworks such as JAX \cite{jax2018github} and MuJoCo JaX (MJX) \cite{zakka2025mujoco}. These tools enable large-scale on-device parallelization and just-in-time compilation through the simulation loop. Most notably, recent toolboxes such as MuJoCo Playground \cite{zakka2025mujoco} are capable of running thousands of MuJoCo-native simulations simultaneously on consumer GPUs, vastly decreasing training times and bolstering a community shift away from CPU-based, small-scale parallelism. 

\footnotetext[1]{\moplayground: \href{https://dynamicmobility.github.io/moplayground/}{https://dynamicmobility.github.io/moplayground/} \label{moplayground}}

\revisediii{However, existing MORL methods have not fully leveraged these benefits of hardware acceleration.}
\revised{Past approaches largely use CPU-based simulation with minimal concurrency}, resulting in training times that are an order of magnitude slower than state-of-the-art single-objective RL. This critical computational gap has fundamentally hindered MORL's application to complex, high-dimensional robot morphologies and settings involving more than a few objectives.

\revisediii{To remove this barrier, this work introduces 
two GPU-native MORL components: an open-source toolbox and a hypernetwork-based algorithm that maps arbitrary desired trade-offs to policy network parameters.}
Our contributions are summarized in Fig. \ref{fig:teaser} and are as follows:
\begin{enumerate}
    \item \morlax: a \revised{JAX-based GPU-compatible \revised{algorithm} for scalable MORL that integrates multi-objective optimization with JAX’s vectorized computation to achieve \speedup~speedups over CPU-based MORL baselines and up to 19x speedups over GPU-based MORL baselines while achieving improved (larger) hypervolumes.}
    \item \moplayground: an open-source toolbox$^{\ref{moplayground}}$ featuring \morlax~and a modernized set of multi-objective MJX environments for systematic benchmarking.


    \item \revisedii{Approximating a 6-objective Pareto set of a humanoid robot (BRUCE) in simulation in $\sim$2 hours, showcasing \moplayground's versatility and performance.}
\end{enumerate}

\section{Background}
\label{sec:background}
\revised{Our work sits at the intersection of MORL and hardware-accelerated computation. We first review the Multi-Objective Markov Decision Process (MOMDP) framework, then survey existing MORL algorithms and toolboxes, and finally introduce \moplayground's parallelized computational paradigm.}

\subsection{Multi-Objective Markov Decision Processes}
\label{sec:morl-theory}
\revised{We adopt the MOMDP framework for multi-objective sequential decision-making} \cite{hayes2022practical}. As with a Markov Decision Process (MDP), a MOMDP is a tuple $(\mathcal{S}, \mathcal{A}, \mathcal{P}, \bm{R}, \gamma)$, where $\mathcal{S}$ is a set of states, $\mathcal{A}$ is a set of actions, $\mathcal{P}(s' \mid s, a)$ is the \revised{probability of transitioning from state $s \in \mathcal{S}$ to state $s' \in \mathcal{S}$ given action $a \in \mathcal{A}$},
and $\gamma \in [0,1)$ is a discount factor. The key distinction from ordinary MDPs lies in the vector-valued reward $\bm{R}: \mathcal{S} \times \mathcal{A} \to \R^m$, \revised{describing $m$ objectives}. \revised{The value function is consequently vector-valued, $\bm{V}_{\pi} \in \R^m$, giving the expected cumulative return per objective:}
\begin{align}
    \bm{V}_{\pi}(s) = \mathbb{E}_{\pi} \left[ \sum_{t = 0}^{\infty} \gamma^t \bm{R}(s_t, a_t) \mid a_t \sim \pi(s_t)\revised{, s_0 = s} \right] \in \R^m.
    \label{eq:vector-value}
\end{align}

We seek to maximize the expected returns over the initial state distribution $\mathcal{D}_{s_0}$:
\begin{align}
\bm{J}_{\pi} = \mathbb{E}_{s_0 \sim \mathcal{D}_{s_0}}\left[\bm{V}_{\pi}(s_0) \right] \in \R^m,
\label{eq:vector-summed-value}
\end{align}
with $\bm{J}_{\pi}$ being shorthand for $\bm{J}(\pi) \in \R^m$.

In most scenarios, no single policy can simultaneously maximize all components of $\bm{J}_{\pi}$ \revised{due to trade-offs among }
the objectives. Thus, the goal is not to find a single optimal policy, but rather to find the set of Pareto-optimal policies, which optimally trade off among the objectives.

Pareto optimality is mathematically quantified using the notion of \textit{Pareto dominance}: a policy $\pi'$ dominates a policy $\pi$ if $\bm{J}_{\pi'} \geq \bm{J}_{\pi}$, where the comparison is taken elementwise, 
and $\bm{J}_{\pi'} \neq \bm{J}_{\pi}$
for at least one component. 
A policy $\pi^*$ is then defined as \textit{Pareto optimal} if it is not dominated by any other policy $\pi'$. 
The set of all such optimal policies forms the \textit{Pareto set}, and their corresponding objectives form the \textit{Pareto front} in the objective space. While early research focused on analyzing the existence of stationary Pareto-optimal policies and solving small-scale MOMDPs via value iteration or linear programming \cite{hayes2022practical}, the challenge for complex robotic systems lies in efficiently approximating the Pareto set with high granularity. Towards this end, several MORL algorithms have been proposed, which we will discuss next.

\subsection{Multi-Objective Reinforcement Learning for Robotics}
Modern MORL approaches focus on generating high-quality approximations of the Pareto set in high-dimensional state and action spaces. One of the first such works, Prediction-Guided MORL (PG-MORL) \cite{xu2020prediction}, developed an evolutionary algorithm that utilizes a prediction model to choose promising optimization directions, resulting in approximate Pareto sets that maximize hypervolume (i.e., the volume of objective space dominated by the discovered solutions) and density (i.e., the degree to which these solutions are evenly distributed). This prior work also established a set of baseline MuJoCo control tasks to benchmark future MORL work \cite{xu2020prediction}, which was later fully standardized by MO-Gym, a toolbox for benchmarking MORL algorithms \cite{felten_toolkit_2023}. Notably, a major disadvantage of PG-MORL is that each policy \revised{in} the Pareto set is represented by a separate neural network, thus requiring thousands of neural networks in cases with 3 or more objectives. This not only affects PG-MORL's learning complexity, but also decreases the granularity of the approximate Pareto set, as only a finite number of policies can be learned.

Subsequent works \cite{hypermorl, liu2025paretosetlearningmultiobjective} address this problem by representing the approximate Pareto \revisediii{set of policy network parameters using} a hypernetwork, a neural network that outputs network \revisediii{parameters} conditioned on a context vector \cite{ha2017hypernetworks}. Namely, HYPER-MORL~\cite{hypermorl} \revised{utilizes the trade-off between objectives as the context vector} 
\revised{of} a hypernetwork \revised{that} outputs the \revisediii{parameters of the} actor and critic networks in the Pareto set, vastly decreasing the number of learnable parameters whilst generating a detailed, continuous Pareto set approximation. However, these approaches utilize small-scale CPU-based parallelization and are thus impacted by long training times, especially in cases with more than two objectives. 

Several other approaches, e.g. AMOR~\cite{alegre2025amor} and PD-MORL~\cite{basaklar2023pd}, learn a single context-conditioned policy to approximate the Pareto set. Of particular interest, AMOR leverages GPU-accelerated parallelization through Isaac Gym. However, AMOR requires 5 days of wall-clock time on an NVIDIA RTX 4090 GPU to approximate a seven-objective Pareto set, which is significantly longer than HYPER-MORL or PG-MORL, albeit with more objectives. \revised{One potential reason for this long training time} is that AMOR requires a large actor network to learn complex mappings between conflicting objectives and desirable actions. This is one of our motivations for leveraging hypernetworks in our approach.


\revised{As a result of long training times, outdated benchmarks, and a lack of open-source integration with modern parallelization, MORL algorithms suffer from substantially prolonged
development pipelines and debugging cycles,
hindering their application 
towards robotics.}

\subsection{GPU-Acceleration for Reinforcement Learning}
Unlike in MORL, \revisediii{which often relies on small-scale CPU-based parallelism,} GPU-accelerated physics simulation is well established in single-objective RL. For example, prior work has demonstrated that a robust single-objective quadruped locomotion policy can be trained within minutes \cite{rudin2022learning}. This massive parallelization alleviates the data-collection bottleneck in on-policy RL and enables the practical deployment of algorithms such as Proximal Policy Optimization (PPO) \cite{schulman2017proximal}. In particular, by drastically shortening training time, GPU-accelerated RL makes it substantially easier to debug and shape reward functions through repeated train–evaluate–modify cycles.

Concurrently, several open-source toolboxes have been 
developed to streamline GPU-accelerated RL. 
One such toolbox, Brax \cite{brax2021github}, introduces a differentiable and parallelizable physics engine built in JAX. Brax became the underlying physics engine and RL framework for MuJoCo Playground, a toolbox that streamlines open-source GPU-based simulation, training, and sim-to-real transfer of several robotic platforms \cite{zakka2025mujoco}. Offering the familiarity and excellent physics accuracy of MuJoCo with proven single-objective RL algorithms like PPO \cite{schulman2017proximal} and Soft Actor-Critic \cite{haarnoja2018soft}, MuJoCo Playground has already spurred many follow-on works \cite{liu2025opt2skill, janwani2025navigait}.

Despite the wide-spread success of GPU-Accelerated RL for robotics, these approaches \revised{are single-objective in nature, and thus cannot flexibly balance multiple objectives in multi-objective problems.}
Some works have attempted to address this lack of tunability, particularly through learning from diverse demonstrations and smoothly blending between reference motions depending on the task specification \cite{liao2025beyondmimic}. Yet, \revised{demonstration collection can be challenging} 
and may not be dynamically feasible for certain robotic platforms.

\subsection{Toolboxes for Multi-Objective Optimization in Robotics}
\begin{table}[t!]
\vspace{2mm}
\centering
\begin{tabular}{l@{\hspace{0.5mm}}ccccc}
\hline
\textbf{Framework} & \begin{tabular}[c]{@{}c@{}}\textbf{\revisedii{Task}}\\ \textbf{\revisedii{Realistic}}\end{tabular} & \textbf{GPU} & \begin{tabular}[c]{@{}c@{}}\textbf{\revisedii{Multiple}}\\ \textbf{\revisedii{Objectives}}\end{tabular} & \textbf{Ref.} \\
\hline
Gymnasium               & \xmark  & \xmark  & \xmark  & \cite{brockman2016gym,towers2024gymnasium} \\
MO-Gymnasium            & \xmark  & \xmark  & \cmark   & \cite{felten_toolkit_2023} \\
MuJoCo Playground       & \cmark  & \cmark  & \xmark  & \cite{zakka2025mujoco} \\
MO-Playground           & \cmark  & \cmark  & \cmark  & Ours \\
\hline
\end{tabular}
\caption{A comparison of open-source RL toolboxes for robotics. \revisedii{MuJoCo Playground and \moplayground~provide task-realistic simulated environments modeled after existing robotic platforms.}
}
\label{tab:framework_comparison}
\end{table}

\begin{figure*}[t!]
    \centering
    \includegraphics[width=\linewidth]{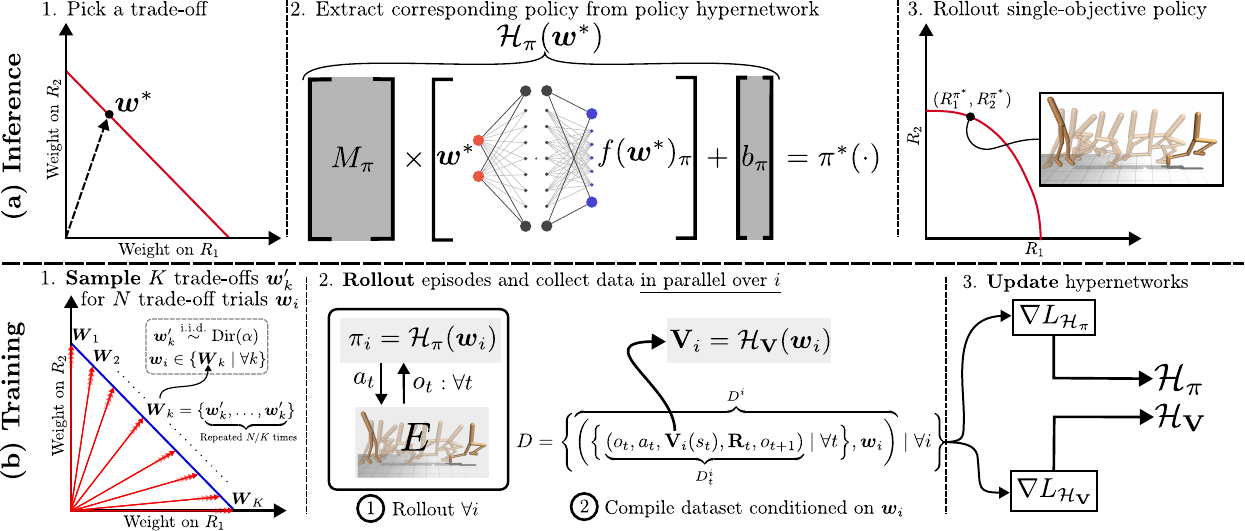}
    \caption{\morlax~Architecture. (a) showcases evaluating the policy hypernetwork $\mathcal{H}_\pi$ by picking a trade-off $\bm{w}$, evaluating $\mathcal{H}_\pi$, and rolling out the extracted policy in the environment. (b) showcases how \morlax~trains $\mathcal{H}_\pi$ via a parallelized \textit{sample}-\textit{rollout}-\textit{update} procedure.}
    \label{fig:architecture}
  \vspace{-3mm}
\end{figure*}
The OpenAI Gym (now called Gymnasium \cite{brockman2016gym}), is likely the most foundational toolbox for RL in robotics. This toolbox standardized the form factor used in RL environments, e.g. the format of reset and step functions to begin episodes and progress through the system dynamics. This toolbox focused primarily on CPU-based single-objective problems, and was soon followed by the Multi-Objective Gym (MO-Gym) \cite{felten_toolkit_2023}. This toolkit paired existing MORL algorithms with MuJoCo tasks contributed by PG-MORL and Gym, among other environments not related to continuous control. However, neither of these toolkits leverages GPU-based parallelization to speed up the simulation loop.  
Adjacent to the MORL space, the evolutionary Multi-Objective Optimization toolbox EvoX \cite{evox} contains a suite of evolutionary single and multi-objective algorithms and interfaces with the aforementioned control tasks. In a recent update, it also interfaces with MuJoCo Playground. While this brings GPU acceleration to evolutionary algorithms in EvoX, the multi-objective algorithms and environment benchmarks still rely on CPU-based computation, limiting parallelism and speed compared to our GPU-based approach. Table \ref{tab:framework_comparison} compares \revised{these} MORL and RL toolboxes, \revised{showcasing} the specific contributions of our work, \moplayground.

\section{\moplayground: GPU-Accelerated MORL}
\label{sec:morlax}


The \moplayground~framework has two major components: (1) \morlax, a multi-objective, highly parallelized actor-critic algorithm that  learns a continuous Pareto set representation, and (2) a suite of multi-objective continuous control benchmarks that build upon prior GPU-accelerated simulation \cite{zakka2025mujoco}. With a simple \texttt{pip install moplayground}, our work enables users to approximate high-quality Pareto sets in drastically shorter times than previous approaches \cite{alegre2025amor, hypermorl}, as shown in Sec. \ref{sec:comparison-results}. 


\subsection{Problem Setup}
As illustrated in Fig. \ref{fig:architecture}, \morlax~learns parameter-efficient Pareto set approximations using hypernetworks; the input to the hypernetwork is a \emph{trade-off vector} $\bm{w} \in \Delta^{m-1}$ on the $(m-1)$-simplex, and the output is the corresponding Pareto-optimal policy \revisediii{network} parameters. The trade-off vector specifies an objective prioritization via the convex combination $\bm{w}^\top\bm{R}$, which \morlax~interprets as an \textit{optimization direction} along which to extend the Pareto front. This linear scalarization is convenient for value-based algorithms like \morlax: it induces a total order over otherwise unsortable vector returns, enabling the discovery of convex Pareto fronts $\mathcal{F}$:
\begin{align}
\mathcal{F} = \{\bm{J}_\pi \mid \exists \bm{w} \textrm{ s.t. } \bm{J}_{\pi} \cdot \bm{w} \geq \bm{J}_{\pi'}\cdot \bm{w}, ~\forall \pi'\}.
\end{align}

\subsection{Hypernetwork Architecture}
\label{sec:architecture}
\morlax~jointly trains two hypernetworks, an actor hypernetwork and a critic hypernetwork, respectively:
\begin{align}
    \mathcal{H}_\pi: \Delta^{m-1} \rightarrow \Theta_{\pi}, \qquad \mathcal{H}_{\bm{V}}: \Delta^{m-1} \rightarrow \Theta_{\bm{V}},
\end{align}
\revisediii{which learn mappings between $\bm{w} \in \Delta^{m-1} \in \mathbb{R}^m$ and respective actor and critic network parameters}, with parameter spaces denoted as $\Theta_{\pi}$ and $\Theta_{\bm{V}}$. We represent both hypernetworks in an affine form. The actor hypernetwork (critic defined analogously) is modeled as: 
\begin{align}
     \mathcal{H}_\pi(\bm{w}) = M_\pi f_\pi(\bm{w}) + b_\pi,
\end{align}
where $f_\pi : \Delta^{m-1} \rightarrow \R^F$ is a nonlinear function that maps trade-off $\bm{w}$ to an $F$-dimensional feature vector. This feature representation is then transformed into the actor parameter space $\Theta_\pi$ via matrix $M_\pi \in \mathbb{R}^{|\Theta_\pi| \times F}$ and column vector $b_\pi \in \mathbb{R}^{|\Theta_\pi|}$. This hypernetwork representation \revisediii{of the mapping between trade-off vector and policy network parameters} is well-studied, as it induces a low-rank parameter manifold \cite{hypermorl}, thereby reducing the problem dimensionality and enabling efficient representations for Pareto set policies.

While $\mathcal{H}_\pi$ outputs an ordinary stochastic policy network that maps observations to a distribution of actions, commonly seen in PPO \cite{schulman2017proximal}, $\mathcal{H}_{\bm{V}}$ outputs a \textit{vector-valued} $\bm{V}_{\pi}$ as defined in \eqref{eq:vector-value}, where $\pi$ is drawn from $\mathcal{H}_\pi$ for some trade-off $\bm{w}$. The use of these hypernetworks is explained further below.

\begin{figure*}[t!] 
  \centering
  \includegraphics[width=\linewidth]{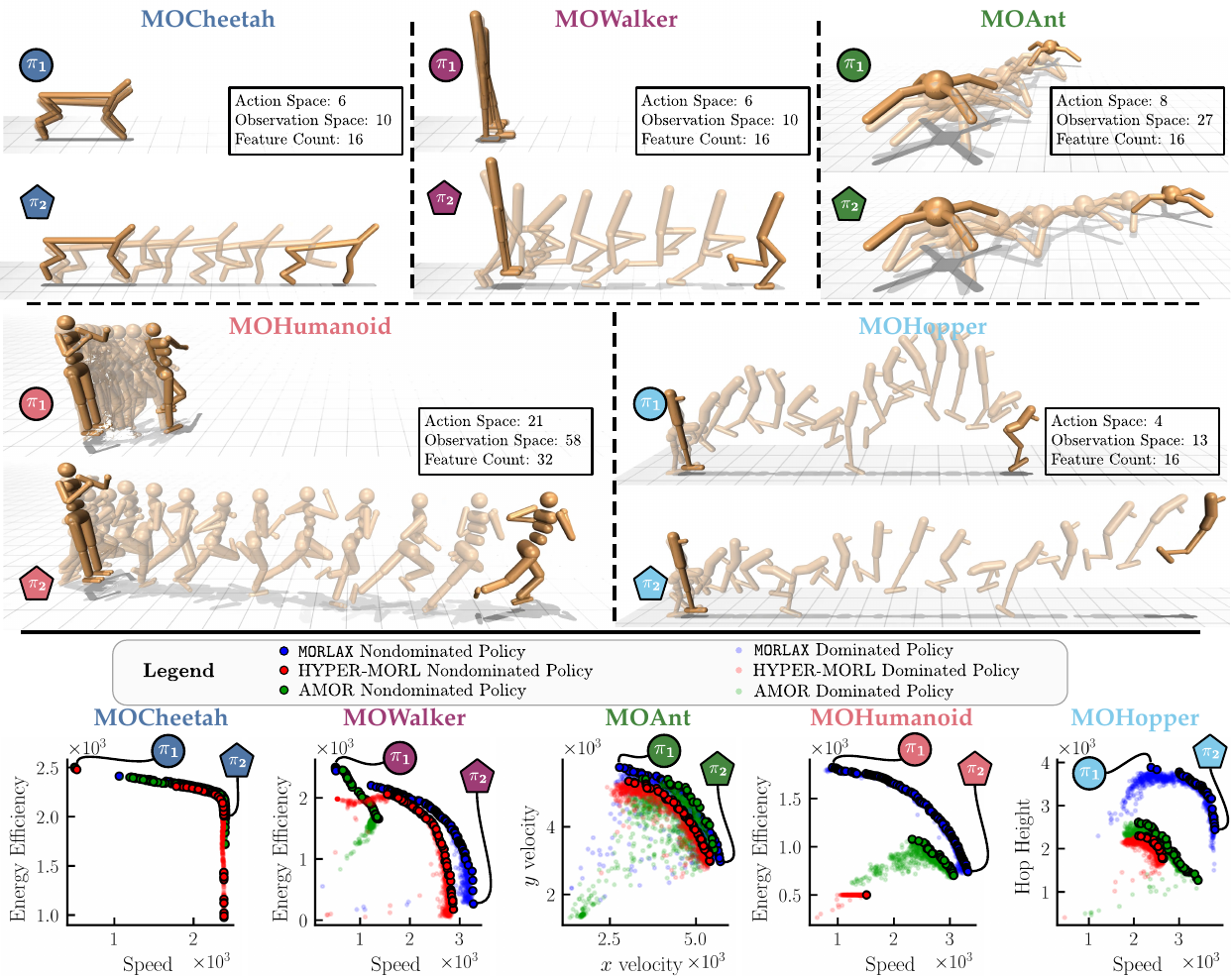}
  \caption{\morlax~finds Pareto sets that achieve better-performing, diverse behavior \revisedii{compared to the baselines, HYPER-MORL \cite{hypermorl} and AMOR \cite{alegre2025amor} while being significantly faster than both: \speedup~speed-up from HYPER-MORL and up to 19x from AMOR}. To visualize each environment's objectives, Pareto fronts along with composite figures of a single \morlax~policy hypernetwork are shown across extreme input trade-offs (e.g. maximum speed vs. maximum energy efficiency in Cheetah). The displayed trajectories are generated via policies highlighted on the Pareto Fronts.}
  \label{fig:comparison-results}
  \vspace{-4mm}
\end{figure*}

\subsection{The \morlax~Algorithm}
Beyond simply moving simulation onto hardware accelerators, \morlax~is explicitly structured to efficiently update its hypernetworks by \revisedii{leveraging} massive parallelization and automatic differentiation via JAX \cite{jax2018github}, achieving high utilization and sample throughput on accelerator devices such as GPUs. 
We separate the algorithm into an initialization phase and three training phases: sampling, rollout, and update. Figure \ref{fig:architecture}b illustrates each phase in detail.

\newsubsec{Algorithm Initialization} \morlax~is 
initialized for a robotic environment $E$ with observation space $\mathcal{O}$ and action space $\mathcal{A}$. The environment is instantiated as $N$ parallel environments, allowing data collection to be vectorized across CPU or GPU workers. The reward returned across environments is a vector-valued signal $\bm{R}: \mathcal{S}\times\mathcal{A} \rightarrow \mathbb{R}^m$ for $m$ objectives. Lastly, we initialize $\mathcal{H}_\pi$ and $\mathcal{H}_{\bm{V}}$ with random weights.

\newsubsec{Sampling Phase}
\label{sec:sampling}
At the start of each algorithm iteration, \morlax~constructs a set of $K$ trade-off vectors,
\begin{align}
\bm{W'} = \{\bm{w}'_1, \dots, \bm{w}'_K\}, \qquad \bm{w}'_i \in \Delta^{m-1},
\end{align}
by 
drawing $K$ i.i.d. samples from a symmetric Dirichlet distribution with concentration parameters $\alpha_i = 1$, for $i = 1, \dots, m$, such 
that each $\bm{w}'_i$ is sampled uniformly 
over the simplex. Importantly, $K > 1$ is an integer factor of $N$, so that $K$ batches of repeated trade-offs are constructed and cleanly parallelized over $N$ parallel environment instances, enabling many trials of each stochastic policy to be generated in the rollout phase
to explore multiple trade-offs. We denote this set of \textit{repeated} $\bm{w}'_i$ as: 
\begin{align*}
\{\bm{w}_i\}_{i=1}^N = \{\bm{W}_i\}_{i=1}^K =\{\underbrace{\bm{w}'_1, \hdots, \bm{w}'_1}_{\textrm{$N / K$ times}}, \hdots, \underbrace{\bm{w}'_K, \hdots ,\bm{w}'_K}_{\bm{W}_K}\}.
\end{align*}
While $K = N$ does not affect \morlax's iteration time, it can affect learning speed and potential; we empirically observe that sparse sampling routines (a few, large clusters of trade-offs) often perform better than dense sampling routines (thousands of independent samples). \revisediii{We hypothesize this stems from reduced gradient variance from averaging over multiple rollouts per trade-off, leading to more stable updates}. To further exploit this finding, \morlax~provides an option to sample $K-\kappa$ clusters of trade-offs, and designate $\kappa$ trade-offs to represent the extreme endpoints within the simplex, such as $[1, 0]$ and $[0,1]$ for $m = 2$. When $\kappa > 0$, the extreme trade-off vectors serve to extend the Pareto set to include even the most extreme policies, e.g. energy efficient policies that remain completely still. Naturally, $\kappa$ scales with the number of objectives and must satisfy $K \geq \kappa$. We reserve $K, \kappa$ as hyperparameters 
\revisediii{and recommend $K \geq 8$ and $\kappa = m$. Please see the appendix on the project website$^\textrm{\ref{moplayground}}$ for more details and an ablation study.}

\newsubsec{Rollout Phase} 
In standard PPO, the rollout phase involves selecting a single policy $\pi_\theta$ that interacts with $N$ parallel environments to collect transitions of the form:
\begin{align}
\mathcal{D}^{(i)}_{\textrm{single objective}}
=
\left\{\left(s_t^{(i)}, a_t^{(i)}, R_t^{(i)}, s_{t+1}^{(i)}\right)\right\},
\end{align}
where $s_t^{(i)} \in \mathcal{S}$, $a_t^{(i)} \in \mathcal{A}$, and $R_t^{(i)} \in \R$ denote sampled states, actions, and rewards with $i$ indexing the parallel environment and $t$ indexing the timestep within rollout horizon $T$. These transitions are then used to compute advantages and update the actor-critic networks.

\morlax~mirrors this 
structure, but replaces PPO's actor-critic networks with a \emph{family} of $N$ (possibly repeated) stochastic actors, each with a corresponding critic. Explicitly, the policy and corresponding value families are denoted: 
\begin{align}
    \{\pi_i\} &:= \{ \mathcal{H}_{\pi}(\bm{w}_i) \mid i = 1,\dots,N\}, \\ \{\bm{V}_i\} &:= \{ \mathcal{H}_{\bm{V}}(\bm{w}_i) \mid i = 1,\dots,N\}.
\end{align}

Then, for each policy, \morlax~collects the corresponding dataset via policy interaction 
with the environment $E$ across the $N$ parallel environments and evaluates $\bm{V}_i$ at each state. The environment rollouts produce datasets $\mathcal{D}^{(i)}$ conditioned on trade-offs $\bm{w}_i$, which are aggregated to form a consolidated dataset over all $N$ parallel instances:
\begin{align}
\mathcal{D} = \{(\mathcal{D}^{(i)}, \bm{w}_i) \mid i = 1,\dots,N\}.
\end{align} 
It is important to note that when $K \neq N$, there are repeated trade-offs and thus \textit{repeated policies} are evaluated. However, because $\mathcal{H}_\pi$ provides a stochastic policy, the collected data will have some variance, especially in the early stages of training, leading to more stable learning.

\newsubsec{Update Phase} \morlax~performs 
separate updates on $\mathcal{H}_\pi$ and $\mathcal{H}_{\bm{V}}$ using the aggregated dataset $\mathcal{D}$. This decoupled optimization allows the two hypernetworks to learn at different rates, which is well established to improve stability and performance in traditional actor-critic methods  \cite{konda1999actor}. 

We adopt a multi-objective extension of the Proximal Policy Optimization (PPO) framework following Xu et al.~\cite{xu2020prediction}. To compute policy updates, we first estimate a vector-valued advantage
$\bm{A}_t \in \mathbb{R}^m$, which estimates the quality of a given action relative to the average action taken in the state, by applying Generalized Advantage Estimation (GAE)
independently to each objective. The resulting advantages are normalized across
the batch, following common PPO practice~\cite{brax2021github}, using the scalarized advantage $A_t^i = \bm{w}_i^\top \bm{A}(s_t^i, a_t^i)$.

Using this scalarized advantage, \morlax~minimizes the unclipped policy and value losses given by:
\begin{align}
L_{\mathcal{H}_\pi}
&=
-\mathbb{E}_{(s_t^i,a_t^i,\bm{w}_i) \sim D}
\left[
r_t(\theta_{i})\, A_t^i
\right],
\\[6pt]
L_{\mathcal{H}_{\bm{V}}}
&=
\mathbb{E}_{(s_t^i,\bm{w}_i) \sim D}
\left[
\left\|
\bm{V}_{\phi_i}(s_t^i) - \hat{\bm{V}}^i_t
\right\|_2^2
\right], \text{ where}
\end{align}
\begin{align}
r_t(\theta_{i})
=
\frac{\pi_{\theta_{i}}(a_t^i \mid s_t^i)}
     {\pi_{\theta_{i},\mathrm{old}}(a_t^i \mid s_t^i)},
~\theta_{i} = \mathcal{H}_\pi(\bm{w}_i), ~\phi_i = \mathcal{H}_{\bm{V}}(\bm{w}_i),
\end{align}
and where $r_t(\theta_{i})$ denotes the policy probability ratio of the policy parameters $\theta_i = \mathcal{H}_{\pi}(\bm{w}_i)$ associated with $\bm{w}_i$, 
$\phi_{i} = \mathcal{H}_{\bm{V}}(\bm{w}_i)$ denotes the vector-valued critic network parameters associated with $\bm{w}_i$, and lastly, $\hat{\bm{V}}^i_t$ denotes the vector-valued value targets computed from rollout returns using GAE. The expectations are taken over all time indices (i.e., $t = 0, \dots T$) and every rollout 
in the dataset (i.e., $i = 1, \dots, N$).

In practice, we optimize the policy hypernetwork using the clipped PPO surrogate objective, which replaces the policy ratio $r_t(\theta)$ with a clipped version to limit excessively large updates. These modifications improve training stability
and are consistent with standard PPO implementations~\cite{brax2021github}.

\subsection{Multi-Objective Environments in \moplayground}

\label{sec:moplayground}

As illustrated in Fig. \ref{fig:comparison-results}, our toolbox includes 5 classic DeepMind control benchmark environments \cite{tunyasuvunakool2020dm_control}: Cheetah, Walker, Ant, Humanoid, and Hopper, initially established in the multi-objective realm by Xu et al.~\cite{xu2020prediction}. We update each of these environments to reflect recent changes in XML structure related to GPU-optimization \cite{zakka2025mujoco} and MuJoCo updates \cite{todorov2012mujoco}. Unique to our package and to facilitate development, all of our environments feature a swappable backend (\texttt{numpy} and \texttt{jax.numpy}) for fast CPU environment evaluation and highly parallelized evaluation during GPU training. 
\begin{table}[t!]
\vspace{0.7em}
\centering
\revisedii{\begin{tabular}{l l !{\color{black!30}\vrule} ccc}
\toprule
\textbf{Environment} & \textbf{Metric} & \textbf{\morlax} & \textbf{AMOR} & \textbf{HYPER-MORL} \\
\midrule
\multirow{4}{*}{Cheetah}
  & Hv & $\bm{5.75}${\tiny$\bm{\times 10^6}$}       & $5.74${\tiny$\times 10^6$}      & $5.58${\tiny$\times 10^6$}     \\
  & $S$                       & $0.008$              & $\bm{0.006}$     & $0.09$       \\
  & $T_{\textrm{base}}$       & 144.0 s  & \textbf{53.9 s}   & 4955.0 s   \\
  & $T_{\textrm{max}}$        & 365.4 s          & 512.0 s  & 4955.0 s   \\
\midrule
\multirow{4}{*}{Walker}
  & Hv & $\bm{6.48}${\tiny$\times 10^6$}      & $3.07${\tiny$\times 10^6$}        & $5.28${\tiny$\times 10^6$}     \\
  & $S$                       & 0.014                & \textbf{0.013}        & 0.015         \\
  & $T_{\textrm{base}}$       & \textbf{142.6 s}  & 1420.3 s        & 4454.4 s   \\
  & $T_{\textrm{max}}$        & 512.2 s                & 1420.3 s        & 8075.8 s   \\
\midrule
\multirow{4}{*}{Ant}
  & Hv & $\bm{3.05}${\tiny$\bm{\times 10^7}$}       & $2.92${\tiny$\times 10^7$}        & $2.70${\tiny$\times 10^7$}     \\
  & $S$                       & \textbf{0.015}                & 0.030       & 0.023         \\
  & $T_{\textrm{base}}$       & \textbf{506.0 s}  & 1771.7 s        & 13338.6 s  \\
  & $T_{\textrm{max}}$        & 1141.7 s               & 2083.6 s        & 13338.6 s  \\
\midrule
\multirow{4}{*}{Hopper}
  & Hv & $\bm{1.48}${\tiny$\bm{\times 10^7}$}       & $8.14${\tiny$\times 10^6$}        & $5.97${\tiny$\times 10^6$}      \\
  & $S$                       & \textbf{0.020}             & 0.021        & 0.052         \\
  & $T_{\textrm{base}}$       & \textbf{53.2 s}   & 1012.9 s        & 12728.2 s  \\
  & $T_{\textrm{max}}$        & 1860.0 s                & 1012.9 s        & 12728.2 s  \\
\midrule
\multirow{4}{*}{Humanoid}
  & Hv  & $\bm{5.45}${\tiny$\bm{\times 10^6}$}        & $3.55${\tiny$\times 10^6$}        & $7.43${\tiny$\times 10^5$}         \\
  & $S$                       & 0.010                & 0.103        & \textbf{0.00}         \\
  & $T_{\textrm{base}}$       & \textbf{92.4 s}   & $1364.6$ s        & 25050.0 s         \\
  & $T_{\textrm{max}}$        & 1866.3 s                & 5213.9 s        & 25050.0 s         \\
\bottomrule
\end{tabular}}
\caption{\revisedii{Comparative results between \morlax, AMOR \cite{alegre2025amor}, and HYPER-MORL \cite{hypermorl}. For each environment, we report the maximum hypervolume attained (Hv), the spacing of solutions on the learned Pareto front ($S$), and the times to achieve the baseline and maximum hypervolumes ($T_{\textrm{base}}$, $T_{\textrm{max}}$). Across all environments, \morlax~achieves larger hypervolumes and maintains similar spacing relative to both HYPER-MORL and AMOR. For all environments except the Cheetah, \morlax~is significantly faster.}}
\label{tab:comparison-table}
\vspace{-4mm}
\end{table}

\subsection{Key Adaptations to HyperMORL}
\label{sec:adaptations}
\begin{figure*}[th!]
    \centering
    \includegraphics[width=\linewidth]{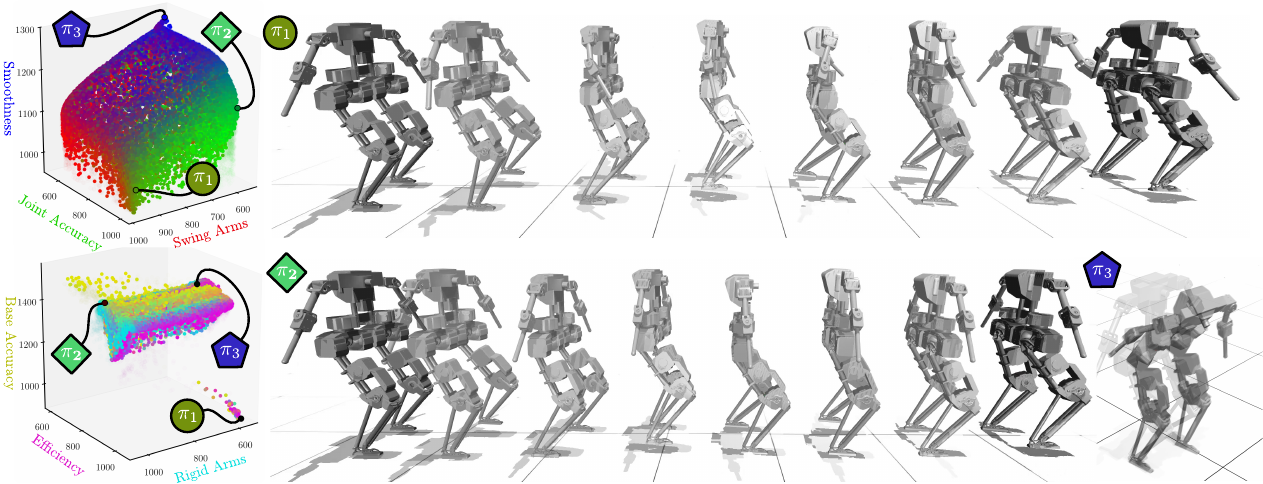}
    \caption{We project the learned six-dimensional Pareto front representing 30,720 policies tracking a moderate, forward velocity onto two 3D plots, each with different objectives. The colors represent trade-off vectors, and are blended according to the color-coded objectives along each axis. $\pi_1$ represents a policy with arm swinging, $\pi_2$ represents a policy with rigid arms, and $\pi_3$ represents a policy that maximizes smoothness. One interesting finding is that $\pi_1$ walks faster and has higher efficiency, likely an emergent benefit of arm-swinging.}
    \label{fig:bruce-pareto}
    \vspace{-3mm}
\end{figure*}
GPU parallelization results in non-sequential network training updates. 
Thus, to maintain HYPER-MORL's training stability~\cite{hypermorl},
two key adaptations are needed. \revisedii{First, HYPER-MORL uses a single hypernetwork with outputs split into actor and critic parameters, coupling their gradients through a shared hypernetwork backbone. In contrast, \morlax, which receives larger and more diverse batches of data via massive parallelization, decouples this structure into separate actor and critic hypernetworks (Sec. \ref{sec:architecture}). This allows each network to be optimized independently, removing interference
\cite{garcin2025studying}, \cite{konda1999actor}}.

\revisedii{Second, we propose a clustered sampling strategy for selecting trade-off vectors (Sec. \ref{sec:sampling}): repeating trade-offs across many parallel environments and pinning a subset of extreme trade-offs in the simplex's corners. This improves gradient stability while extending the Pareto set to its boundary.}
\revised{We conduct an ablation study which supports these adaptations, and is located in the appendix on the \moplayground~website$^\textrm{\ref{moplayground}}$.} 
Furthermore, \morlax~does not require a warm-up stage due to increased data-collection enabled by massive parallelization.
\revisedii{We also adopt standard PPO practices, such as utilizing $\tanh$ activation functions for the actor to account for actuator limits, and distinguishing between terminations and truncations in value estimation.}

\section{Results}
\label{sec:results}
To validate our approach, we compare the results of \morlax~and baselines HYPER-MORL \cite{hypermorl} and AMOR \cite{alegre2025amor} on each \moplayground~environment. Further, to demonstrate scalability, we show a practical application of \morlax~on \revisedii{a high-fidelity simulation of the BRUCE humanoid robot}.

\subsection{Comparison with MORL \revisedii{Baselines}}
\vspace{-2mm}
\label{sec:comparison-results}

\newsubsec{Implementation Details}
We train \revisediii{\morlax~and baseline algorithms HYPER-MORL (CPU-based) and AMOR (GPU-based) on the same environments and tasks on the same desktop machine, which uses} an NVIDIA RTX A4000 GPU. \revisediii{We use comparable numbers of learnable parameters} across \revisediii{all algorithms} within each environment, \revised{allocating more for the Humanoid and BRUCE due to their more complex task specification and dynamics.} 
\revisedii{See \moplayground's website$^{\ref{moplayground}}$ for implementation parameters and usage documentation.}

\newsubsec{Runtime comparison} \revised{We first compare \morlax's speed-up over} \revisedii{HYPER-MORL and AMOR} for identical \moplayground~environments--\revisediii{using the same desktop machine for fair comparison}. \revised{To compare runtimes, we run each algorithm to convergence, take the smallest final hypervolume across the three as a baseline $\textrm{Hv}_{\textrm{base}}$, and measure the time $T_{\textrm{base}}$ each algorithm takes to reach $\textrm{Hv}_{\textrm{base}}$.}
Table \ref{tab:comparison-table} reports \revised{these} results: a massive speedup (of $T_{\textrm{base}}$) delivered by \moplayground, \speedup~over HYPER-MORL and up to 19x over AMOR. 
\revisedii{Interestingly, AMOR is faster than \morlax~for the Cheetah, which has simple tasks (speed vs. energy efficiency) and dynamics (statically stable). This highlights that \morlax~is well-suited for complex multi-objective problems, achieving higher quality Pareto fronts faster than AMOR as task and dynamic complexity increase.} 
\revisedii{To study the impact of hardware acceleration, we also report environment steps per second for the Cheetah for both hypernetwork-based algorithms: 82,972 steps/sec for \morlax~and 3,631 steps/sec for HYPER-MORL, a 22.8x
improvement. This is below \morlax's total 34.4x speed-up over HYPER-MORL, signaling that \morlax's algorithmic adaptations account for the remainder of the speedup.}

\newsubsec{\revisedii{Pareto set} comparison} \revisedii{We report both the hypervolume and spacing \cite{pymoo} of the approximate Pareto sets to assess their quality.} As observed in Table \ref{tab:comparison-table}, \revisedii{\morlax~learns Pareto fronts with superior hypervolume and similar spacing across all five benchmark environments.}
\revisedii{Specifically, improvements are modest on Cheetah, Walker, and Ant, but substantially larger on Hopper and Humanoid. Notably, the Hopper's speed and height objectives 
require complex control strategies to optimize, in contrast to energy efficiency, which is optimized by standing still. Unlike the Hopper and other environments, the Humanoid features 3D, unstable dynamics, making even simple objectives, such as standing (or efficiency), nontrivial. Together, this demonstrates \morlax's applicability to problems with complex objectives and dynamics.}


\subsection{Custom Robotic Example: BRUCE Humanoid}
\label{sec:bruce-results}

We showcase the new capabilities and substantial speed-up afforded by \morlax~through its application to a \revisedii{practical}, highly multi-objective robotics problem: humanoid locomotion. Prior MORL methods \revisedii{require several days to train on similar tasks~
\cite{alegre2025amor}}, exemplifying their infeasibility for many practical scenarios. \revisedii{In this work, we apply \moplayground~to a high-fidelity simulation of the BRUCE~humanoid robot \cite{westwoodrobotics_bruce}, using identical domain randomization and controller structure as \cite{janwani2025navigait}, which achieved hardware deployment.} We straightforwardly \revised{instantiate this environment as} a child class of the \texttt{MultiObjective} \revised{class} provided by \moplayground, extending standards \revised{set} by MuJoCo Playground \cite{zakka2025mujoco}.

\newsec{Robot Model}                                                  \revised{We extend the BRUCE model from \cite{janwani2025navigait} to include upper limbs, yielding 16 actuated joints (3 per hip, 1 per knee, 1 per ankle pitch, 2 per shoulder, 1 per elbow) within a 20-DoF local configuration; the remaining 4 DoFs come from the passive ankle rolls and the four bar linkage constraints. We refer to \cite{janwani2025navigait} for further modeling details.}



\newsec{NaviGait Controller}
\revised{We build on the \textsc{NaviGait} controller introduced in~\cite{janwani2025navigait}. \textsc{NaviGait} utilizes RL to select and track motions in a gait library generated via trajectory optimization. This strategy yields visually appealing behavior, as both nominal and recovery motions can be sourced from the gait library.} \revised{Notably, \textsc{NaviGait}'s original approach \cite{janwani2025navigait} utilized single objective RL, thus requiring nontrivial reward tuning. Instead, we use \moplayground~to explore Pareto-optimal trade-offs across objectives and even discover new behaviors, such as coordinated upper-body motion that complements the underlying locomotion. To facilitate \moplayground~integration, we alter the \textsc{NaviGait} policy to directly output joint commands, rather than the residual corrections used in \cite{janwani2025navigait}.}



\newsec{Reward Design}
Six separate reward functions were constructed for MORL with BRUCE: base tracking, joint tracking, arm swinging, arm rigidness, energy efficiency, and smoothness. Explicit reward definitions along with their respective numerical constants can be found \revised{in \moplayground's repository linked on the project website$^\textrm{\ref{moplayground}}$.} 

\newsec{Results}
In Fig. \ref{fig:bruce-pareto}, we showcase the results of extending \moplayground~to \revised{BRUCE}. Across the aforementioned six objectives, \morlax~finds continuous, highly diverse Pareto sets in $\sim$2 hours and 11 minutes. \revisedii{Though demonstrated on a different task and robot platform, this is a significant speedup compared to the 5-day training time reported in AMOR \cite{alegre2025amor}}. The resulting policies are capable of following forward, lateral and angular yaw command velocities in \revised{domain randomized environments}. \revised{Fig. \ref{fig:bruce-pareto} shows three examples of the diverse} behaviors contained within the approximate Pareto set: swinging arms ($\pi_1$), rigid arms ($\pi_2$), and maximum smoothness ($\pi_3$). One interesting finding is that policies that swing their arms walk faster with higher efficiency, as seen when comparing $\pi_1$ to $\pi_2$. 
We show only two projections of this high-dimensional Pareto front for a moderate, forward velocity command \revised{and include more visualizations and videos on the project website$^{\ref{moplayground}}$}.





\section{Conclusion}

\revised{\moplayground~is an open-source, GPU-native MORL toolbox for robotics that enables fast, high-quality Pareto set approximation over many objectives. 
However, it shares some common limitations with existing MORL approaches. First is the assumption that objectives are known \textit{a priori}, when many human-specified objectives, like \textit{naturalness}, are difficult to write down mathematically. \morlax~also uses a linear scalarization function, like most MORL algorithms for continuous control \cite{xu2020prediction, hypermorl, alegre2025amor}, limiting Pareto front approximations to be convex.} Future extensions of our work include human-in-the-loop optimization to guide multi-objective reward learning, alongside applications in human-centered robotics such as 
lower-limb exoskeletons.





\bibliographystyle{IEEEtran}
\bibliography{references}

\end{document}